\documentclass[journal,twoside,web]{IEEEtran}

\usepackage{amssymb}

\usepackage[figuresright]{rotating}
\usepackage{amsmath}
\usepackage[inline]{enumitem}
\usepackage{xcolor}

\usepackage[hidelinks]{hyperref}

\usepackage{verbatim}
\usepackage{float}

\begin{document}

\title{
	{\fontsize{18.6}{23}\selectfont Don't Think It Twice: Exploit Shift Invariance for Efficient Online Streaming Inference of CNNs}
}

\author{
\IEEEauthorblockN {
   Christodoulos Kechris,
   Jonathan Dan,     
   Jose Miranda,
   David Atienza
}

\thanks{This research was supported in part by the Swiss National Science Foundation Sinergia grant 193813: "PEDESITE -
Personalized Detection of Epileptic Seizure in the Internet of Things (IoT) Era", and the Wyss Center for Bio and Neuro Engineering: Lighthouse Noninvasive Neuromodulation of Subcortical Structures.}
\thanks{All authors are affiliated with the Embedded Systems Laboratory (ESL), EPFL, Switzerland.}
\thanks{*Corresponding author C.K. e-mail: christodoulos.kechris@epfl.ch}
}%%% author

\maketitle

\begin{abstract}
Deep learning time-series processing often relies on convolutional neural networks with overlapping windows. This overlap allows the network to produce an output faster than the window length. However, it introduces additional computations. This work explores the potential to optimize computational efficiency during inference by exploiting convolution's shift-invariance properties to skip the calculation of layer activations between successive overlapping windows. Although convolutions are shift-invariant, zero-padding and pooling operations, widely used in such networks, are not efficient and complicate efficient streaming inference. We introduce StreamiNNC, a strategy to deploy Convolutional Neural Networks for online streaming inference. We explore the adverse effects of zero padding and pooling on the accuracy of streaming inference, deriving theoretical error upper bounds for pooling during streaming. We address these limitations by proposing signal padding and pooling alignment and provide guidelines for designing and deploying models for StreamiNNC. We validate our method in simulated data and on three real-world biomedical signal processing applications. StreamiNNC achieves a low deviation between streaming output and normal inference for all three networks (2.03 - 3.55\% NRMSE). This work demonstrates that it is possible to linearly speed up the inference of streaming CNNs processing overlapping windows, negating the additional computation typically incurred by overlapping windows.
\end{abstract}

\section{Introduction}
In many DL time-series inference applications, such as robotics or healthcare, the model's output is required when a new sample is acquired, an inference scheme referred to as streaming inference. In such an environment, computation optimization methods that rely on batch-parallelization \cite{mittal2019survey}, \cite{wang2019benchmarking} are not applicable. Other optimization strategies have been proposed to reduce inference computations and boost efficiency, such as pruning \cite{li2020penni}, dynamic pruning \cite{shen2020fractional} and early exiting \cite{li2023predictive}. 

Time-series DL models often process overlapping data windows, guaranteeing frequent output and robustness on sample border effects \cite{zanghieri2019robust}, \cite{kechris2024kid}, \cite{shahbazinia2024resource}, \cite{reiss2019deep}. This overlap introduces additional computational overhead, and a considerable portion of the input information is repeated between successive windows. Shift-invariance is the ability to maintain the same representations when the input is temporally translated. It can be used to mitigate this additional overhead during online inference. Convolution provides a natural way to introduce shift invariance to a DL model \cite{bruintjes2023affects}. 

Kondratyuk et al. \cite{kondratyuk2021movinets} and Lin et al. \cite{lin2019tsm} proposed specific streaming Convolutional Neural Network (CNN) architectures for processing videos. For one-dimensional time series, Khandelwal et al. \cite{khandelwal2021efficient} presented a real-time inference scheme for CNNs. Their method is tested on CNNs of limited depth and fixed kernel size without pooling operations. These systems exploit the temporal translational invariance of the convolution operation, allowing them to skip computations between successive windows and update only the required layer outputs. 

These methods cannot be generalized for performing streaming inference with any temporal CNN architecture. Additional operations, such as pooling, padding and dense layer, are generally used, which are not translation invariant.

Pooling reduces the dimensionality of intermediate representations temporal axis \cite{zanghieri2019robust}, \cite{kechris2024kid}, \cite{shahbazinia2024resource}. \cite{azulay2019deep} empirically investigated the connection between Nyquist's sampling theorem and the lack of translation invariance due to pooling. A low-pass filter on the pooling filter was proposed by Zhang \cite{zhang2019making} as a potential solution to limit the effects of aliasing. Other works have also proposed similar anti-aliasing filters to mitigate the effect of pooling \cite{chaman2021truly}, \cite{zou2023delving}.

CNNs also use zero-padding \cite{o2015introduction}. This introduces positional information in the learned representations, breaking their translation invariance \cite{kayhan2020translation} and further complicating the deployment of pre-trained CNNs as streaming models.

In this work, we propose exploiting common information in successive windows to reduce computations during inference. Specifically, we focus on CNNs, exploiting convolution's inherent translation invariance properties \cite{von2003modern}. We investigate the two main CNN components that break translation invariance: padding and pooling. Based on this exploration, we derive StreamiNNC, a strategy for adapting any pre-trained CNN for online streaming inference. We evaluate our proposed method on three real-world biomedical streaming applications.

We present the following novel contributions:
\begin{itemize}
    \item We introduce StreamiNNC, a scheme for efficient CNN inference in streaming mode requiring minimal changes to an original pre-trained CNN.
    \item We investigate the effect of zero-padding on the accuracy of StreamiNNC inference and compare it to the alternative of signal-padding.
    \item We derive a signal-padding training strategy with minimal changes to the original CNN and the training routine. 
    \item We provide a theoretical explanation of the translation invariance properties of pooling which, so far, have only been investigated empirically.
\end{itemize}
Our code is available here: \url{https://github.com/esl-epfl/streaminnc}

\section{Preliminaries}
Denote a real-time domain signal $x(t): \mathbb{R} \rightarrow \mathbb{R}, t \in \mathbb{R}$. Without loss of generality, we consider a single-channel signal, although our analysis can be expanded to the multi-channel case: $\boldsymbol{x}(t): \mathbb{R} \rightarrow \mathbb{R}^{N_{channels}}, t \in \mathbb{R}$. The signal is sampled, with a sampling period $T_s$, into its discrete representation, $x[i] = x(i \cdot T_s), i \in \mathbb{N}$ and then windowed with windows of length $L$ and step $S$ samples. 

We denote a vector from successive sampled points $\{x(n_1 \cdot T_s), x((n_1 + 1) \cdot T_s), \dots, x(n_2 T_s)\}$, where $n_1 < n_2$ and $n_1,n_2 \in \mathbb{N}$ as: $\boldsymbol{x}_{n_1:n_2} = [x[n_1], ..., x[n_2]]$. Then the $i-th$ window, from $t_i = i \cdot T_s \cdot S$ to $t_i + L \cdot T_s$, is the vector $\boldsymbol{x}_i = \boldsymbol{x}_{i : i + L}$. If $S < L$, then between two successive windows, $\boldsymbol{x}_{i : i + L}$ and $\boldsymbol{x}_{i + S : i + S + L}$, there is overlap, that is, there are $L-S$ common samples $\boldsymbol{x}_{i + S : i + L}$. We assume that $L - S > 0$, $S > 0$, and to simplify our derivations, we consider $L$ to be divisible by $S$.

Let a real time-domain signal $x(t)$, $f: \mathbb{R} \rightarrow \mathbb{R}^M$ operating on $x$ and $T_{\Delta t}[\cdot]$ is the operation of translating a signal in time by $\Delta t$, i.e. $T_{\Delta t}[x](t) = x(t + \Delta t)$. $f$ is temporally translation invariant if:

\begin{equation} \label{eq:translation_invariance}
    f(T_{\Delta t}[x]) = T_{\Delta t}[f(x)], \quad \forall \Delta t \in \mathbb{R}
\end{equation}

For on-line inference, a deep CNN, $f(\cdot)$, serially processes the windows $\boldsymbol{x}_i$ as soon as they are available. The CNN comprises a feature extractor $h(\cdot)$ and a feature classifier, $g(\cdot)$, $f = h \circ g$. $h$ is composed of a series of $N$ convolutional blocks, $h_j$: $h = h_0 \circ \dots \circ h_j \circ \dots \circ h_{N-1}$. Each convolutional block can contain convolution layers, activations, batch normalization layers, and pooling (subsampling) operations. $g$ is typically a series of fully connected layers followed by non-linear activations. 

For the entire feature extractor $h$ to be shiftable, all layers $h_i, i \in [0, N - 1]$ have to be shiftable. Layers that process a single activation point are trivially shiftable, e.g., for ReLU $y[i] = max(x[i], 0), \forall x[i] \in \boldsymbol{x}$. Shiftability in layers processing a group of points, e.g. convolutions or pooling, is not trivial and requires elaboration. The classifier, $g$, is usually comprised of fully connected layers. Hence, they are not inherently shift invariant, and we do not consider $g$ in StreamiNNC.

\subsection{Convolution and Temporal Translation Invariance}
For a linear kernel $w(t) \in \mathbb{R}$ the convolution $(w \ast x)(t): \mathbb{R} \rightarrow \mathbb{R}$ is: $y(t) = (w \ast x)(t) = \int w(\tau)x(t - \tau)d\tau$. Shifting the input signal $x(t)$ by $\Delta t$ results in the same output but shifted by $\Delta t$ as well, satisfying eq. \ref{eq:translation_invariance}: $T_{\Delta t}[y](t) = \int w(\tau)x(t + \Delta t - \tau)d\tau$.

In the time-finite, discrete case, the situation is similar, yet with some nuances. The discrete convolution output, $\boldsymbol{y}_i \in \mathbb{R}^{L-M + 1}$, of $\boldsymbol{x}_i$ and kernel $\boldsymbol{w} \in \mathbb{R}^M$ is: $y_i[n] = \sum w[m]x_i[n-m]$. Consider the convolution output for the next window $\boldsymbol{x}_{i + 1}$ where $\boldsymbol{x}_i$ is shifted by $S$ samples. Except for the boundary samples, the output is again equivalent to the previous output, just shifted by $S$ samples: $y_{i + 1}[n] = y_i[n + S] = \sum w[m]x_i[n + S - m]$.

However, in practice, the input window is usually padded with $M - 1$ zeros such that the output remains the same size as the input: $\boldsymbol{y} \in \mathbb{R}^L$. Then $\boldsymbol{y}_i = \boldsymbol{w} \ast [\boldsymbol{0} \quad \boldsymbol{x}_i \quad \boldsymbol{0}]$ and the shift equivalence no longer holds, since the output border elements are affected by the padding zeros.

\section{Methods}
\begin{figure*}[h]
    \centering
    \includegraphics[width=0.7\linewidth]{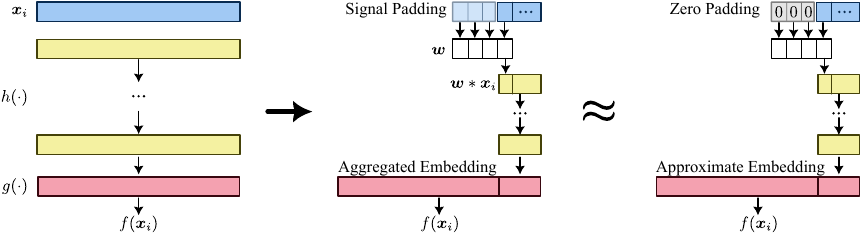}
    \caption{\textbf{Left: Full Inference.} CNN, $f$, processing a window $\boldsymbol{x}_i$. \textbf{Middle: Streaming Inference.} Only the new information is processed by $f$, and part of the inputs and activations are stored and retrieved to be used as padding for the next window. All intermediate embeddings are stored in the aggregated embedding. If the network has been trained with Signal Padding, then the aggregated embedding is equivalent to the full inference embedding. \textbf{Right: Approximate Streaming Inference.} Just like in streaming inference, we only process the newest samples. Here, previous inputs/activations are not stored, and zero-padding is used instead as an approximation. The resulting intermediate embeddings are aggregated into an approximate embedding.}
    \label{fig:streaming_inference_demo}
\end{figure*}

Given a pre-trained CNN $f = h \circ g$, StreamiNNC optimizes online inference by operating $h$ in streaming mode, applying the minimum amount of changes to the original network (Figure \ref{fig:streaming_inference_demo}). We achieve this by replacing window-wide convolutions with convolutions processing only the new information. Depending on the architecture of $f$, past information may be stored for exact streaming inference or discarded for approximate streaming. Additionally, StreamiNNC may require weight retraining to guarantee shiftability. In the rest of this section, we explore the factors determining these design choices.

\subsection{Padding}
Padding is necessary to maintain a deep network structure without degrading the activation to a single value. However, zero padding destroys the convolution's shift-ability. It is also problematic in terms of representing an infinite signal. In signals like images, the signal is confined within the limited pixels of the image, with the rest being considered just zero values. However, a time-infinite signal $x(t)$ is not only constrained within the limits of the $\boldsymbol{x}_i$. 

We address these zero-padding limitations with Signal Padding (Figure \ref{fig:streaming_inference_demo}), inspired by the Stream Buffer \cite{kondratyuk2021movinets}. In Signal Padding, the input of each convolutional layer is padded with values of the previous window, i.e. for a window $\boldsymbol{x}_i$ its signal-padded equivalent is: $[\boldsymbol{x}_{i - M: i} \quad \boldsymbol{x}_i]$. Then the output of the first convolution layer with weights $w^1$ is $\boldsymbol{y^1}_i = \boldsymbol{w^1} \ast [\boldsymbol{x}_{i - M: i} \quad \boldsymbol{x}_i]$. Since we are dealing with online inference we have adopted a causal convolution scheme \cite{bai2018empirical}. In general, for the convolutional layer at depth $d$:
\begin{equation} \label{eq:signal_padding_convolutions}
    \boldsymbol{y}^d_i = \boldsymbol{w}^d \ast [\boldsymbol{y}^{d-1}_{i - M : i} \quad \boldsymbol{y}^{d-1}_i]    
\end{equation} The padding values of the activations, $\boldsymbol{y}^{d-1}_{i - M: i}$, need to be buffered between successive window inferences. 

\subsection{Pooling}

\begin{figure}[h]
    \centering
    \includegraphics[width=0.9\linewidth]{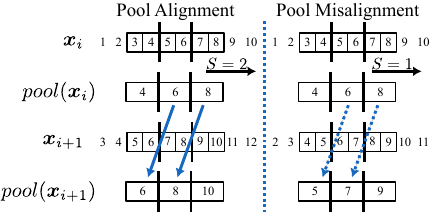}
    \caption{Illustration of shift-invariance of the pooling operation. \textbf{Left:} During the previous window, $i - 1$, the sequence $[1 \cdots 6]$ is processed. Then the window moves by a step of $S = 2$ samples, window $i$, processing samples $[3 \cdots 8]$ and similarly for the $i + 1$ window. The input is passed through Max Pooling with a pooling window size $L_p = 2$. $S$ and $L_p$ are aligned, hence $pool(\boldsymbol{x}_{i + 1})$ can be partially estimated from the elements of $pool(\boldsymbol{x}_{i})$ (blue arrows). \textbf{Right:} $S$ and $L_p$ are misaligned, and the pooling operation is not shiftable. Shifting the elements of $pool(x_i)$ to partially estimate $pool(x_{i + 1})$ can only be an approximation.}
    \label{fig:pooling_demo}
\end{figure}

In general, pooling is not invariant to temporal translation. It can be if we constrain the window step $S$ to be a multiple of the pooling window length $L_p$, Figure \ref{fig:pooling_demo}. This constraint has to be guaranteed for all pooling operations in the network. In the general case, however, pooling can be approximately shift invariant. We will now investigate the shift approximation of pooling by deriving upper error bounds for approximating a pooling operation as shiftable. 

Let $p: \mathbb{R}^{L_p} \rightarrow \mathbb{R}$ be a pooling operation $y_i = p(\boldsymbol{x}_{i : i + L_p})$. $p(\cdot)$ takes as an input a vector of $L_p$ samples, performs an operation and outputs one scalar value as the result. For example, for Max Pooling, $p(\boldsymbol{x}_{i : i + L_p}) = max(\boldsymbol{x}_{i : i + L_p})$. 

Now $\boldsymbol{x}$, the input to a pooling operation, is sampled from a continuous signal $x(t)$, bandwidth limited to $f_{max}$, that is, for its Fourier transform, $X(f)$, it holds that $X(f) = 0, \forall f > f_{max}$. Letting $A = sup|x(t)|$, and following Bernstein's inequality \cite{pinsky2023introduction}, the absolute first time derivative of $x(t)$ is bound by: $|x'(t)| \leq 2 \cdot \pi \cdot A \cdot f_{max}$, where $|x^{\prime}|$ can reach the maximum $2 \cdot \pi \cdot A \cdot f_{max}$ only when $x(t)$ contains a single oscillation at $f_{max}$, $x(t) = A cos(2 \pi f_{max} t + \phi)$. Since $x[n]$ is sampled from $x(t)$, $\frac{x[i] - x[i - 1]}{T_s} \approx x'(t)$, and

\begin{equation} \label{eq:consequtive_samples_bound}
    |x[i] - x[i - 1]| \leq  2 \cdot \pi \cdot A \cdot f_{max} T_s = 2 \cdot \pi \cdot A \cdot \frac{f_{max}}{f_s}
\end{equation}
Note that due to the Nyquist theorem, $f_{max} < f_s / 2$, hence in the worst case scenario: $|x[i] - x[i - 1]| <  \pi \cdot A$ and $|x[i] - x[i - 1]| \leq 2A < \pi \cdot A$. So the upper bound saturates when $f_{max} > f_s / \pi$ at $2A$.

From eq. \ref{eq:consequtive_samples_bound} we can derive the upper error bounds for approximating a pooling operation by shifting. Consider two samples $x_i$ and $x_j$ with $j - i = m > 0$, $|x[i] - x[j] | \leq (m - 1) \cdot 2 \cdot \pi \cdot A \cdot \frac{f_{max}}{f_s}$. Here we have considered the worst-case scenario in which $x$ maintains the maximum rate for all samples from $x[i]$ to $x[j]$. Additionally, we assume the worst-case scenario where the pooling windows are misaligned such that they share only one common time-point sample $x[i]$.

\textbf{Pooling}
In this case, the pooling operation is defined as $y^p_i = p(\boldsymbol{x}_{i : i + L_p}) = x_i$. In the worst case scenario, the pooling window is misaligned such that the shifting output is $y^p_{i + L_p} = p(\boldsymbol{x}_{i + L_p : i + 2L_p}) = x_{i + L_p}$. Then

\begin{equation} \label{eq:pooling_approximation_error_upper_bound}
    \mathbb{E}[|y^p_{i + S} - y^p_{i}|] = \mathbb{E}[| x_{i + L_p} - x_i |] \leq (L_p - 1) \cdot 2 \cdot \pi \cdot A \cdot \frac{f_{max}}{f_s}
\end{equation}

\textbf{Max Pooling}
For Max Pooling, $p(\boldsymbol{x}_{i : i + L_p}) = max(\boldsymbol{x}_{i : i + L_p})$, and similarly to the basic pooling case:

\begin{align} \label{eq:max_pooling_approximation_error_upper_bound}
    \mathbb{E}[|y^p[i + S] - y^p[i]|] \leq (L_p - 1) \cdot 2 \cdot \pi \cdot A \cdot \frac{f_{max}}{f_s}
\end{align}

\textbf{Average Pooling} 
For average pooling, $p(\boldsymbol{x}_{i : i + L_p}) = \frac{1}{L_p} \sum_{n = i}^{L_p} \boldsymbol{x}_{i : i + L_p}[n]$ and:

\begin{align} \label{eq:avg_pooling_approximation_error_upper_bound}
    \mathbb{E}[|y^p[i + S] - y^p[i]|] \leq (L_p - 1) \cdot 2 \cdot \pi \cdot A \cdot \frac{f_{max}}{f_s}
\end{align}

The following corollaries can be drawn from eq. \ref{eq:pooling_approximation_error_upper_bound}, \ref{eq:max_pooling_approximation_error_upper_bound} and \ref{eq:avg_pooling_approximation_error_upper_bound}:
\begin{enumerate}
    \item Given $S$ - $L_p$ misalignment, $L_p > 1$ and a finite $f_s$, the upper bound cannot guarantee strict equality unless $f_{max} = 0$, i.e. constant input. 
    \item However, the approximation error of the shift can be small enough, given a high enough sampling frequency of the pooling input, $\boldsymbol{x}$, compared to its bandwidth. In this case, pooling does not achieve a high-dimensionality reduction.
    \item In the worst-case scenario, the error can be considerable, for example, the relative error for Max Pooling $\frac{\mathbb{E}[|y^p[i + S] - y^p[i]|]}{A} \leq 2 (L_p - 1), 2 (L_p - 1) > 1$.
\end{enumerate}

To generalize for the entire CNN, $\boldsymbol{x}$ is given as input to $f(\cdot)$, with $x(t)$ band-limited at $f_{max}$. As $\boldsymbol{x}$ traverses the network layers, each layer will affect its spectral content. Linear convolutions may limit $f_{max}$ through linear filtering but, being linear, cannot expand the bandwidth. However, non-linear activations will increase it, introducing additional frequencies higher than the original $f_{max}$ \cite{kechris2024dc}. Potentially $f_{max}$ can be increased close to the Nyquist maximum $f_s / 2$.

Apart from the non-linear activations, the pooling operations themselves are also affecting the upper error bound in two ways:
\begin{enumerate}
    \item The effective sampling frequency is reduced as the original input passes through successive pooling layers.
    \item $f_{max}$ may also be reduced if the input to the pooling contains frequencies higher than the effective Nyquist frequency. Since these cannot be represented, they are discarded.
\end{enumerate}
Overall, the upper bounds of the shift approximation error become increasingly looser for deeper layers of the CNN. Our conclusion is consistent with the empirical observations of \cite{azulay2019deep}. Additionally, these upper bounds provide insights into why anti-aliasing is insufficient even when $f_{max} < f_s / 2$. Recall that for $f_s / \pi < f_{max} < f_s / 2$ the upper error bound saturates at $2 A (L_p - 1)$. If $f_s >> f_{max}$, then we can safely assume that the pooling layer is approximately shiftable.  

\subsection{Streaming and Approximate Streaming Inference}
We now describe streaming inference with shift-able convolutional feature extractors. We make the following assumptions for all layers $h_i$ in $h$ to ensure that $h$ is shiftable:
\begin{enumerate}
    \item The weights have been retrained with signal padding or zero-padding effects are small
    \item Pooling operations are shift-able or approximate shift-able.
\end{enumerate}Once $S$ samples are available they form the sub-window input $\boldsymbol{x_{S_i}}, i \in [0, L/S]$ and are processed by the feature extractor $h$: $E_{S_i} = h(\boldsymbol{x}_{S_i})$. When all $L/S$ sub-windows have been processed, they are aggregated into a single embedding $E_{S_{0:L/S}} = [E_{S_0}, E_{S_1}, \dots, E_{S_{L/S}}]$. $E_{S_{0:L/S}}$ contains the information of the entire window of $L$ samples and is equivalent to the embedding $E_{S_{0:L/S}}$ if $h$ had processed the entire window, $E^{\prime} = h(\boldsymbol{x})$. From there, the classifier $g$ can process this embedding producing its output $y_0 = g(E_{S_{0:L/S}})$. The output is equivalent to processing directly the entire window since the embeddings $E$ and $E^{\prime}$ are equivalent. 

When the next sub-window is processed $E_{S_{1:L/S + 1}} = h(\boldsymbol{x}_{S_{L/S + 1}})$ it is appended to $E$: 
\begin{equation} \label{eq:shifted_embeddings}
    E_{S_{1:L/S + 1}} = [E_{S_1}, E_{S_2}, \dots, E_{S_{L/S}}, E_{S_{L/S + 1}}]
\end{equation} The embeddings $E_{S_1}, E_{S_2}, \dots, E_{S_{L/S}}$ have already been processed, and we do not need to calculate them. We just need an aggregation buffer to hold them in memory. 

Signal padding enables the equivalence $[h(\boldsymbol{x}_{S_0}), \dots, h(\boldsymbol{x}_{S_{L/S}})] = h(\boldsymbol{x})$. However, each convolution requires $M - 1$ samples from the output of the previously processed window to be used as padding, Eq. \ref{eq:signal_padding_convolutions}. These have to be saved into a buffer reserved for each convolutional layer in the feature extractor. Hence, the memory footprint increases since additional space is needed for the buffers of each layer. Additionally, the required read/write operations for these buffers will affect execution time by increasing latency.

If the convolution kernels are small enough, then a small number of padding values are needed. Thus, padding with zeros instead of signal values might be a good enough approximation with respect to inference accuracy. This strategy allows us to avoid the additional buffer and, consequently, the additional memory overhead needed by Signal Padding. Furthermore, this strategy does not require retraining with Signal Padding, and a pre-trained model can be directly deployed for streaming inference. 

\subsection{Training Streaming for Inference}
To guarantee shiftability of the CNN representations during StreamiNNC the pretrained weights of the CNN have to be swapped with signal padding weights, and the CNN needs to be retrained with Signal Padding.

Signal Padding inference relies on sequential processing of temporal data. However, training with sequential execution significantly increases the training time, as there is little data parallelism. In addition, it complicates data shuffling, which might be useful in converging to an optimal solution.

\begin{figure}[h]
    \centering
    \includegraphics[width=0.75\linewidth]{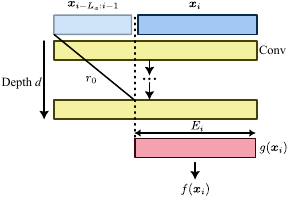}
    \caption{Strategy for training Signal Padding in batch mode. The input window, $\boldsymbol{x}_i$, is extended by $L_a$ samples. $L_a$ is chosen such that at depth $d$ the receptive field of $h$, $r_0$, is smaller than $L_a$. The feature extractor only processes the samples that correspond to the initial window $\boldsymbol{x}_i$.}
    \label{fig:online_training}
\end{figure}

We propose the following Signal Padding training strategy (Figure \ref{fig:online_training}). The feature extractor's, $h$, input window, $\boldsymbol{x}_i$, is extended by appending $L_a$ additional time-points to the input $\boldsymbol{x_{SP}} = [\boldsymbol{x}_{i-L_g : i - 1}, \boldsymbol{x}_i]$, with $\boldsymbol{x}_{i-L_g : i - 1}$ the additional signal samples. $L_a$ has to be selected so that it is at least equal to the receptive field of the deepest convolutional layer in the feature extractor, $r_0$ \cite{araujo2019computing}: $L_a \geq r_o$. $h$ is using zero-padding and the entire network $f$ is trained normally without any further changes.

The output of the feature extractor is then: $h([\boldsymbol{x}_{i-L_g : i - 1}, \boldsymbol{x}_i]) = [E_{i-L_g : i - 1}, E_{i}]$. $E_{i-L_g : i - 1}$ is affected by the zero-padding, while $E_{i}$ is not. Only $E_{i}$ is fed into the classifier $g$, and the network output is $g(E_{i})$. In this way, the network makes an inference only on the original window inputs without the effect of zero padding.

\subsection{Computational Speedup}
\label{sec:methods:expected_computational_speedup}
Under streaming mode, for a new sample $\boldsymbol{x}_{S_{L/S + 1}}$ only $E_{S_{L/S + 1}}$ needs to be computed, eq. \ref{eq:shifted_embeddings}. The rest of the sub-embeddings $E_{S_i}$ have been already calculated in the previous window input and are just restored from the buffer. Hence, in streaming mode, we need $\times L/S$ less operations compared to full inference, leading to $\times (L/S)$ speedup. This speedup estimate ignores the additional overhead from accessing the layer buffers and thus is only accurate for approximate streaming inference.

\section{Experiments}
We validate our theoretical derivations and evaluate our streaming inference methodology by performing experiments on simulated and real-world data. For real-world applications, we consider the following three convolutional networks, processing three different signal modalities.

\textbf{Heart Rate Extraction}. The first CNN, $f_{PPG}$, is inferring heart rate from photoplethysmography signals \cite{kechris2024kid}. The network is evaluated on the PPGDalia dataset \cite{reiss2019deep}, simulating in-the-wild conditions for heart rate monitoring from wearable smartwatches. The signal is sampled at $32 Hz$ and windowed into windows of 8 seconds (256 samples) with a step of 2 seconds (64 samples). The feature extractor of $f_{PPG}$, $h_{PPG}$, consists of three convolutional blocks. Each block contains three convolutional layers with ReLU activation followed by an average pooling operation. All convolutional layers have a large kernel size ($kernel = 5, dilation = 2$) which translates to potentially significant shiftability degradation of $h_{PPG}$ due to the effect of zero-padding, especially in the deeper layers. 

\textbf{Electroencephalography-based Seizure Detection}. The second network, $f_{EEG}$, is performing seizure detection from electroencephalography signals \cite{shahbazinia2024resource} on the Physionet CHB-MIT dataset \cite{shoeb2009application}. The signals are windowed with a window size of 1024 samples and a step of 256 samples. Here, the feature extractor, $h_{EEG}$, comprises three convolutional layers with ReLU activation, followed by batch normalization and max pooling. Here, in contrast to $h_{PPG}$, the kernel size is small ($kernel = 3, dilation = 1$).

\textbf{Wrist acceleration based Seizure Detection}. The last CNN, $f_{ACC}$, is classifying seizures using the acceleration recorded from the patient's wrist \cite{sphar2024epilepsy}. $h_{ACC}$ processes windows of 960 samples with a 160 sample step size and comprises six ReLU convolutional layers ($kernel = 3$) followed by a batch normalization layer. No pooling is utilized here. 

We perform the following experiments.

\textbf{Pooling Error Bounds}. We numerically evaluate our theoretical model from eq. \ref{eq:pooling_approximation_error_upper_bound}, \ref{eq:max_pooling_approximation_error_upper_bound} and \ref{eq:avg_pooling_approximation_error_upper_bound}. We generate example signals and evaluate streaming errors on a single pooling layer when the window step $S$ is not aligned with the pooling window $L_p$. 

Two input cases are considered: a mono-frequency signal, $x_{mono}(t) = cos(2 \pi \cdot f_0 \cdot t)$, and a multi-frequency one $x_{multi}(t) = \sum_{i} cos(2 \pi \cdot i \cdot f_0 \cdot t)$. We sample two 8-second windows from $x_{mono}(t)$ and $x_{multi}(t)$ with an overlap $L - S$: $\boldsymbol{x}_{1_{mono}}, \boldsymbol{x}_{2_{mono}}$ and $\boldsymbol{x}_{1_{multi}}, \boldsymbol{x}_{2_{multi}}$. A swipe over $S$ is performed to estimate the maximum error. For every tested overlap, we calculate the max pooling output for each window and the average and maximum relative errors: $\frac{1}{N}\sum_{i = 0}^{N - S- 1} |\boldsymbol{x}_{1_{m}}[i + S] - \boldsymbol{x}_{2_{m}}[i]|/ A$ and $max_{i \in [0, N - S - 1]} |\boldsymbol{x}_{1_{m}}[i + S] - \boldsymbol{x}_{2_{m}}[i]| / A$ respectively, where $m \in \{mono, multi\}$ and $A = sup|x(t)|$. 

To test the effect of the sampling frequency, we fix the pooling window at 8 seconds and test with sampling frequencies $f_s \in [2^5, ..., 2^{10}] Hz$. Then, we test for pooling window sizes $L_p \in [2^1, ..., 2^{16}]$ fixing $f_s = 256 Hz$.

\textbf{Zero Padding Effect}. We empirically investigate the effect of padding on the translation invariance of $h_{PPG}, h_{EEG}, h_{ACC}$. We set all convolution weights to the same constant value to perform moving averaging and provide as input a constant vector at $1$. Without the effect of zeros in the padding, all convolution output should be 1. The deviation of output samples from 1 indicates the effect of zero padding.

\textbf{Streaming Inference}. We evaluate streaming inference with zero padding on $f_{PPG}, f_{EEG}, f_{ACC}$, using the pre-trained weights from \cite{kechris2024kid}, \cite{shahbazinia2024resource} and \cite{sphar2024epilepsy} and perform inference using StreamiNNC, with exact and approximate streaming. To compare full inference to streaming inference, we compare the outputs of the models between the two modes using Normalised Root Mean Squared Error: $NRMSE(y_{full}, y_{stream}) = \frac{\sqrt{\mathbb{E}[(y_{full}, y_{stream})^2]}}{max(y_{full}) - min(y_{full})} $. $f_{EEG}$ and $f_{ACC}$ are classifiers with two output units, indicating seizure or not-seizure, so we report the NRMSE of the linear activations for each output unit separately, that is, before applying the softmax. We also demonstrate the effect of Window step/ Pooling window misalignment on $f_{PPG}$.

To investigate signal padding, we retrain $f_{PPG}$ with signal padding training. In addition to the $NRMSE(y_{full}, y_{stream})$ we also evaluate its performance as the Mean Absolute Error (MAE) between the model output and the ground truth heart rate \cite{reiss2019deep}, \cite{kechris2024kid}. We also perform partial streaming inference, where only the first three convolutional layers, the least affected by zero-padding, are in stream inference mode, limiting the effect of zero-padding. 

Furthermore, we implemented the $f_{ACC}$ model in C++11 to evaluate the speedup achieved with streaming inference.

\section{Results}
\subsection{Pooling Error Bounds}
\begin{figure}[h]
    \centering
    \includegraphics[width=\linewidth]{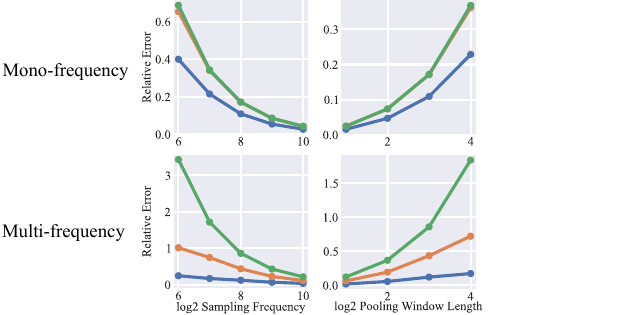}
    \caption{Error introduced due to shifting on non-aligned pooling operations: empirical maximum expected error (blue), empirical maximum error (orange) and derived upper error bounds (green). For the mono-frequency input (top), our bound aligns with the empirical maximum errors. For the multi-frequency input (bottom), the actual empirical error is less than our derived bounds. Nonetheless, our model predicts the behavior of the shift approximation as a function of the pooling window and the sampling frequency.}
    \label{fig:pooling_theoretical_bounds}
\end{figure}

The empirical errors and theoretical upper bounds for the pooling shift approximations are presented in Figure \ref{fig:pooling_theoretical_bounds}. For the mono-frequency input, the empirical maximum error matches our upper bound (top). As expected, for the multi-frequency case (bottom), our error upper bound is larger than the empirical maximum error. In both cases, the empirical average error is lower than our upper bound. Additionally, selecting a small pooling window size or a large enough sampling frequency with respect to the input's bandwidth results in a very small relative error. 

\subsection{Zero padding Effects}
\begin{figure}[h]
    \centering
    \includegraphics[width=\linewidth]{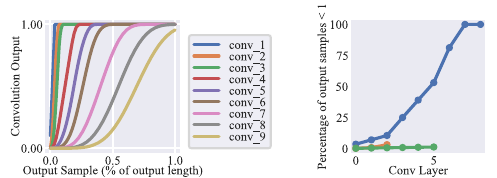}
    \caption{Effect of zero-padding on the convolution activations. \textbf{Left:} Activations of intermediate convolution layers from $h_{PPG}$ with constant inputs at 1 and moving average convolutional weights. The first layers, e.g. first three, show little zero-padding effect, with the majority of the output at 1, in contrast to deeper layers where all points are affected. \textbf{Right:} Percentage of activation points which are less than 1, indicating an effect of the zero-padding for $h_{PPG}$ (blue), $h_{EEG}$ (orange), and $h_{ACC}$ (green).}
    \label{fig:activations_effect_on_padding}
\end{figure}

\begin{figure}[h]
    \centering
    \includegraphics[width=0.9\linewidth]{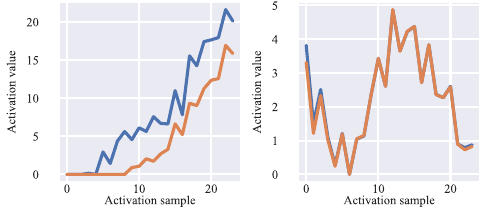}
    \caption{Activations from a representative channel of the last convolutional layer of $h_{PPG}$ with zero padding (left) and signal padding (right) when processing real photoplethysmography data taken from the PPGDalia dataset. The activations of two consecutive windows are presented, with the first window in blue and orange in the second. The activations are temporally aligned such that their values should align if $h_{PPG}$ is shiftable. Zero padding is damaging convolution shiftability causing a difference between re-calculating the activations (orange) and storing them (blue), NRMSE $18.91\%$. In contrast, signal padding allows $h_{PPG}$ to store the activations of the previous window and re-use a part of them for the next input window, NRMSE $2.60\%$.}
    \label{fig:ppg_activations_effect_on_padding}
\end{figure}

The effect of zero-padding values on the activations is presented in Figure \ref{fig:activations_effect_on_padding}. The large kernel size relative to the intermediate representation sizes across the temporal axis causes the network to be considerably affected: after the fifth layer, more than $50\%$ of the activation outputs are affected by the zeros in the padding. This ratio grows to $100\%$ for the last two layers. This is problematic during streaming inference since the shiftability of the convolutional layers is heavily hampered, resulting in a high NRMSE $18.91\%$ (Figure \ref{fig:ppg_activations_effect_on_padding}). Signal padding addresses this issue, reducing NRMSE to $2.60\%$. 

In contrast, for $h_{EEG}$ and $h_{ACC}$, the zeros have a considerably smaller effect. $h_{EEG}$ has a relatively large input (1024 samples), and although it employs pooling, the convolution kernel size and network depth are small enough such that at the last layer, only $3.12\%$ of the activation samples are affected by zeros. In the extreme case, $h_{ACC}$ has a large input (960 samples), a small kernel size of 3 samples and no pooling operations. As such, the zero padding affects $1.25\%$ of the output of the convolution. 

\subsection{Streaming and Approximate Streaming Inference}

The performance of $f_{PPG}$ as a StreamiNNC model is presented in Table \ref{tab:hr_mae}. Streaming inference with zero-padding leads to an increase in the model's inference error (Streaming MAE $4.45 BPM$ vs $3.86 BPM$ for full inference). This error can be reduced by partial streaming (MAE $3.77 BPM$) without retraining the network. Retraining with signal padding also reduces the streaming error (MAE $3.37 BPM$ streaming vs $3.36 BPM$ full). The misalignment of the window step / grouping leads to a significant increase in the inference inaccuracies ($6.73 BPM$). 

\begin{table}[h]
    \centering
    \begin{tabular}{l|c|c}
        Inference & MAE (BPM) & NRMSE (\%) \\
        \hline
        \hline
        \multicolumn{3}{c}{\textbf{Pretrained Zero Padding Weights}} \\
        \hline
        Full  & 3.84 & -\\
        Streaming  & 4.42 & 5.83\\ 
        Partial & 3.77 & 3.02 \\
        Pool Mis. & 6.73 & 7.85\\
        \hline
        \multicolumn{3}{c}{\textbf{Retrained Signal Padding Weights}} \\
        \hline
        Full & 3.36 & - \\
        Streaming & 3.38 & 2.03\\
        \end{tabular}
        \caption{MAE and NRMSE for $f_{PPG}$ for full and streaming inference. }
    \label{tab:hr_mae}
\end{table}

StreamiNNC, without any signal padding retraining, performs satisfyingly for all models (Table \ref{tab:streaming_nrmse}). Especially $f_{EEG}$ and $f_{ACC}$ present small deviations between streaming and full inference, NRMSE between $3.32\%$ and $3.55\%$. $f_{PPG}$ presents the largest deviation (NRMSE $5.83\%$). These findings are aligned with our exploration of the zero-padding effect (Figure \ref{fig:activations_effect_on_padding}). Finally, $f_{ACC}$ presents a satisfyingly small deviation even with approximate StreamiNNC (NRMSE $2.12\%$), indicating the lack for the need of additional buffers.

\begin{table}[h]
    \centering
    \begin{tabular}{l|c|c|c}
        Inference & $f_{PPG}$ & $f_{EEG}$ & $f_{ACC}$   \\
        \hline
        \hline
        Streaming & 5.83 & 3.32 / 3.45 & 3.55 / 3.51\\
        Approx. Streaming & 19.9 & 7.39 / 7.50 & 2.12 / 2.13 \\ 
        \end{tabular}
        \caption{\%NRMSE of streaming and approximate streaming for pre-trained networks using zero-padding. For $f_{EEG}$ and $f_{ACC}$, the NRMSE of both output channels are presented.}
    \label{tab:streaming_nrmse}
\end{table}

\subsection{Streaming Speedup}

\begin{figure}[H]
    \centering
    \includegraphics[width=0.4\linewidth]{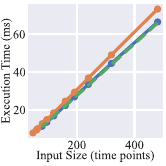}
    \caption{Linear execution time vs input size for approximate streaming inference (blue), streaming inference(orange) and our theoretical estimate (green).}
    \label{fig:speedup_times}
\end{figure}

The inference speedup achieved is presented in Figure \ref{fig:speedup_times}. For the approximate streaming inference, the speedup is $\times (L/S)$, (coefficient $0.13$). In exact streaming inference, an additional overhead is added because of the buffers needed to store embeddings from previous samples. The speedup is thus less than $\times (L/S)$ but still presents a linear behavior (coefficient $0.15$). 

\section{Discussion}
From our theoretical exploration and experiments, we derive the following guidelines for StreamiNNC.

\textbf{Signal vs Zero Padding.} Signal Padding guarantees shiftability of the CNN and hence equivalence between full and streaming inference. However, it requires specialized retraining, which might be impossible, e.g. private dataset, or cost-ineffective. Conversely, a pre-trained CNN can be directly deployed with StreamiNNC, without any changes if the architecture allows it. Our \textit{constant input - moving average filter} method (Experiments - Zero Padding Effect) provides an intuitive way of evaluating an architecture's temporal shiftability. In our experiments we were able to achieve a satisfyingly low error (NRMSE $3.02\%$, Table \ref{tab:hr_mae}) even with a $10.54\%$ effect of zero-padding (Figure \ref{fig:activations_effect_on_padding}).

\textbf{Exact vs Approximate Streaming.} Approximate streaming can be useful for some architectures, e.g. $h_{ACC}$ (NRMSE $2.12 - 2.13\%$, Table \ref{tab:streaming_nrmse}), especially since it does not require any additional buffers for signal padding, reducing the memory footprint of the network. However, the inaccuracies added to the sub-embeddings $E_{S_i}$, due to the zero-padding (Figure \ref{fig:streaming_inference_demo}), can introduce considerable errors. In the worst case scenario, this can render the output useless, e.g. $f_{PPG}$ (NRMSE $19.9\%$ Table \ref{tab:streaming_nrmse}). The effect is dependent on the padding sizes throughout the network, similarly to \textit{Signal vs Zero Padding}, however, our experiments indicate that here the output is more sensitive to the zero effects. 

\textbf{Pooling Alignment.} Ensuring that the window step is aligned with the pooling window size is crucial for guaranteeing model shiftability. Failing to do so can introduce significant errors, e.g. in our case 7.85\% NRMSE for $f_{PPG}$ (Table \ref{tab:hr_mae}). Pooling methods optimized for translation-invariance \cite{zhang2019making} \cite{chaman2021truly}, \cite{zou2023delving} could help reduce this error. This would have to be analysed and compared to the upper error bounds derived in this paper.

\textbf{The Clasifier.} In this work we have only dealt with streaming the feature extractor sub-network, $h$. The classifier, $g$, usually comprises layers lacking the shift-invariant property, e.g. fully connected layers \cite{kechris2024kid}. This can be partially mitigated using a Fully Convolutional Neural Network configuration \cite{long2015fully}, which however would require retraining a new classification sub-network. 

\section{Conclusions}

In this work, we have introduced StreamiNNC, a strategy for operating any pret-rained CNN as an online streaming estimator. We have analyzed the limitations posed by padding and pooling. We have derived theoretical error upper bounds for the shift-invariance of pooling, complementing empirical insights from previous works. Our method allows us to achieve equivalent output as standard CNN inference with minimal required changes to the original CNN. Simultaneously it achieves a linear reduction in required computations, proportional to the window overlap size, addressing the additional computational overhead introduced by the overlap. 

% \clearpage

\bibliographystyle{ieeetr} % We choose the "ieee" reference style
\bibliography{refs} % Entries are in the refs.bib file

\end{document}